\documentclass[lettersize,journal]{IEEEtran}
\usepackage{amsmath,amsfonts}
\usepackage{algorithmic}
\usepackage{algorithm}
\usepackage{array}
\usepackage[caption=false,font=normalsize,labelfont=sf,textfont=sf]{subfig}
\usepackage{textcomp}
\usepackage{stfloats}
\usepackage{url}
\usepackage{verbatim}
\usepackage{graphicx}
\usepackage{cite}
\usepackage{amsmath, amssymb}
\usepackage{multirow}
\usepackage[table,xcdraw]{xcolor}
\usepackage{colortbl}
\usepackage{makecell}
\usepackage{graphicx} 
\usepackage{tabularray}
\usepackage{ragged2e}

\usepackage{array}
\usepackage{multirow}
\usepackage{ragged2e}
\usepackage[font=footnotesize,labelfont=bf,textfont=normal]{caption} 
\begin{document}

\title{SplitFrozen: Split Learning with Device-side Model Frozen for Fine-Tuning LLM on Heterogeneous Resource-Constrained Devices}


\author{
    Jian Ma, 
    Xinchen Lyu, 
    Jun Jiang, 
    Qimei Cui, 
    Haipeng Yao, and 
    Xiaofeng Tao\thanks{
        J. Ma, X. Lyu, Q. Cui and X. Tao are with Beijing University of Posts and Telecommunications, China, and also with Peng Cheng Laboratory, China;
        
        J. Jiang is with Peng Cheng Laboratory, China;
        
        H. Yao is with Beijing University of Posts and Telecommunications, China.
    }
}

\markboth{Journal of \LaTeX\ Class Files,~Vol.~14, No.~8, August~2021}%
{Shell \MakeLowercase{\textit{et al.}}: A Sample Article Using IEEEtran.cls for IEEE Journals}


\maketitle

\begin{abstract}
Fine-tuning large language models (LLMs) on private, on-device data can empower tailored personalized AI agents. However, fine-tuning LLMs on resource-constrained edge devices faces significant challenges, including excessive  computation overhead, device heterogeneity, and data imbalance. This paper proposes SplitFrozen, a split learning framework that enables efficient LLM fine-tuning by strategically freezing device-side model layers while centralizing parameter-efficient fine-tuning on the server. Our framework partitions LLMs into device-side frozen layers and server-side fine-tuning layers, where heterogeneous resource-constrained devices execute only forward propagation. To minimize server-side training costs, we integrate Low-Rank Adaptation (LoRA) into the server-side layers. A pipeline parallelism strategy further optimizes training efficiency by decoupling device-server computations and leveraging decomposed backward propagation. Experiments on GPT-2 with the MRPC, MNLI-matched, and SST-2 datasets demonstrate that SplitFrozen outperforms FedLoRA and SplitLoRA by 69.4\% model accuracy under extremely imbalanced data, 
 while reducing up to 86.8\% device-side computations and 50.2\% total training time. Experiments also validate the scalability of SplitFrozen on content generation task using Llama-3.2 model on GSM8K dataset.
\end{abstract}


\section{Introduction}

The integration of artificial intelligence (AI) and communication has been identified as a key usage scenario for future 6G networks by the International Telecommunication Union (ITU). Large language models (LLMs), such as DeepSeek, SORA, GPT-4o, and Gemini, have demonstrated unprecedented capabilities, and are the underlying AI technologies for the integration of AI and communication in 6G~\cite{ref1}.
Fine-tuning LLMs on private, on-device data can empower tailored personalized AI agents that adapt to individual user contexts. However, the growing complexity of LLMs (often exceeding 400 billion parameters) renders traditional fine-tuning methods computationally prohibitive, particularly for resource-constrained edge devices. 

Parameter-Efficient Fine-Tuning (PEFT) techniques, such as Low-Rank Adaptation (LoRA), enables computation-efficient model fine-tuning by updating only a small subset of model parameters via low-rank matrices. Integrating LoRA for efficient and distributed
model training at the network edge is critical for the
integration of AI and communication in 6G~\cite{ref2}.Existing studies have proposed various approaches for empowering distributed model fine-tuning in 6G, which can be broadly categorized into two frameworks, i.e., FedLoRA~\cite{reffedlora} and SplitLoRA~\cite{refsplitlora}, as summarized in Fig. 1.

\textit{(1) FedLoRA}~\cite{reffedlora} combines federated learning (FL) with LoRA, where devices locally update low-rank matrices via LoRA to be synchronized with a central server. The local LoRA fine-tuning of the full LLM model still exceeds the capabilities of resource-constrained devices.

\textit{(2) SplitLoRA}~\cite{refsplitlora} combines split learning (particularly, Splitfed~\cite{splitfed}) with LoRA to further reduce device-side computation overhead, where the LLM is partitioned into device-side (fine-tuning locally) and server (offloaded to server) portions. Devices fine-tune a subset of layers and transmit intermediate features (i.e., activations) to the server, which sends gradients back. Meanwhile, a local aggregation server is responsible for synchronizing the device-side model portions.

However, existing federated and split learning frameworks, FedLoRA [3] and SplitLoRA [4], still face two challenges. \textit{(1) Device Heterogeneity.} Existing methods enforce uniform computational loads (e.g., full-model fine-tuning in FedLoRA or identical model partitions in SplitLoRA) despite vast disparities in device capabilities  (ranging from high-performance servers to resource-constrained IoT devices).  Fine-tuning (backward via low-rank decomposition) even portions of LLMs would strain devices with limited resources. \textit{(2) Data Imbalance.} Non-IID (non-independent and identically distributed) data distributions would lead to biased model updates, degrading model accuracy in federated and split learning setups.
To the best of our knowledge, efficient fine-tuning LLMs on heterogeneous resource-constrained devices and imbalanced data has yet to be addressed in the literature.

This article proposes SplitFrozen (as shown in Fig. 1(c)), which enables efficient LLM fine-tuning across heterogeneous resource-constrained devices and imbalanced data. Our key insight stems from the observation that pre-trained LLMs exhibit inherent scalability, 
i.e., fine-tuning only a subset of layers (e.g., the last nine layers of GPT-2 on the MRPC dataset) can match the accuracy of full-model 
tuning. Leveraging this property, SplitFrozen strategically freezes device-side model layers, restricting mobile devices to perform only forward propagation, while centralizing backward passes and parameter updates on the server via LoRA. SplitFrozen proposes a three-stage pipeline to efficiently speed up the model fine-tuning process across devices and the server.

The proposed SplitFrozen framework introduces three innovations: \textit{(1) Heterogeneity-aware Partitioning.} Devices deploy varying 
numbers of frozen layers based on their computational capacity, with the server dynamically adjusting its forward depth; \textit{(2) Non-IID Resilience.} Activations from all devices are aggregated and shuffled  at the edge server, mitigating data distribution skew; 
\textit{(3) Pipeline Parallelism.} Overlaps device-side computation, activation transmission, and server fine-tuning to reduce total fine-tuning time. Experiments on GPT-2 with the MRPC, MNLI-matched, and SST-2 datasets demonstrate that SplitFrozen outperforms FedLoRA and SplitLoRA 
by  69.4\% model accuracy under extremely Non-IID data, while reducing up to 86.8\% device-side computations and 50.2\% total training time. SplitFrozen is also shown to scale on content-generation tasks using the Llama-3.2 model on GSM8K dataset.

\begin{figure*}[ht]
  \centering
  \begin{minipage}[t]{0.48\textwidth}
    \centering
    \includegraphics[width=0.9\linewidth]{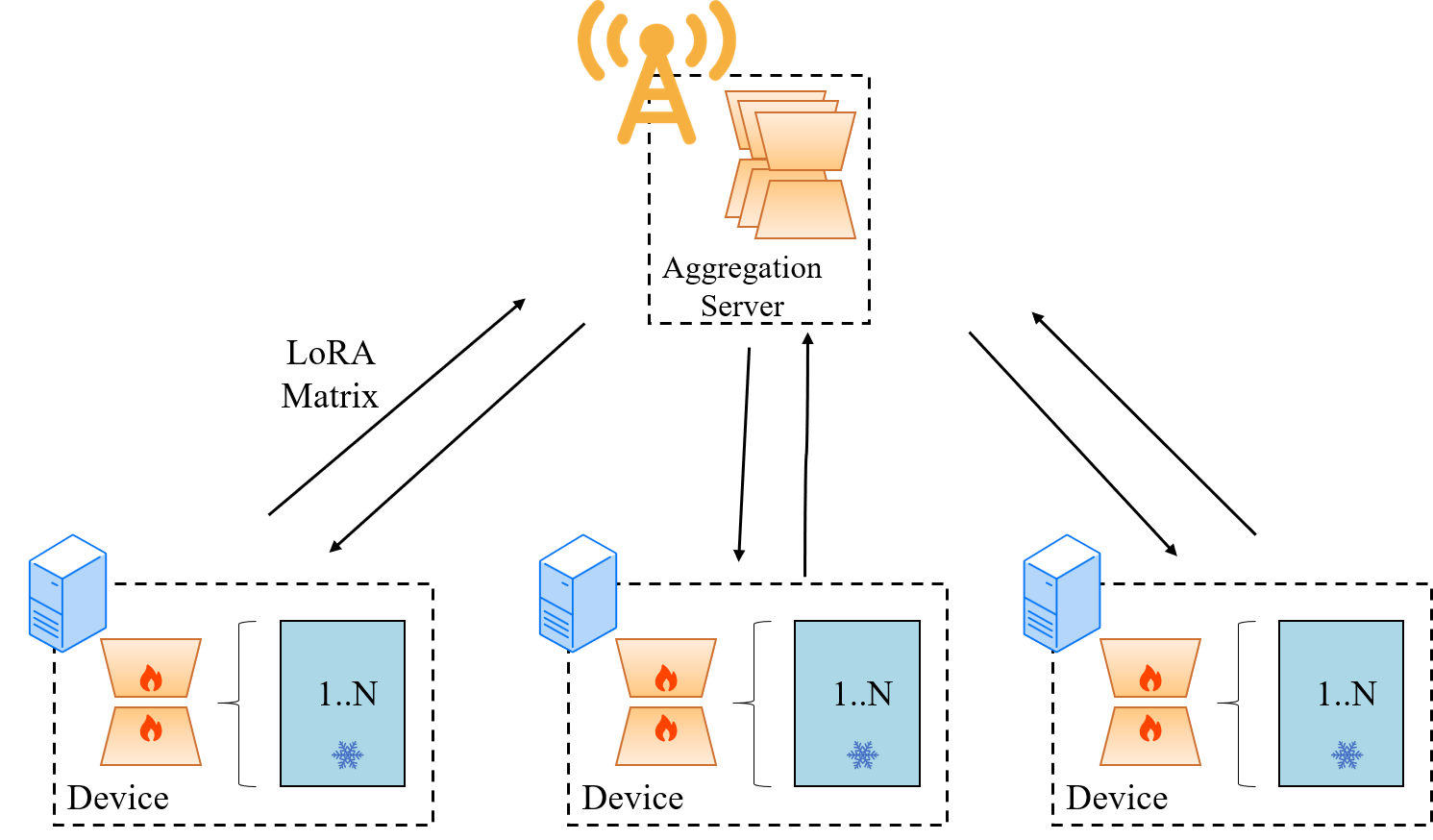}
    \caption*{(a) FedLoRA[3]} 
    \label{fig:fedlora}
  \end{minipage}%
  \hfill
  \begin{minipage}[t]{0.48\textwidth}
    \centering
    \includegraphics[width=0.9\linewidth]{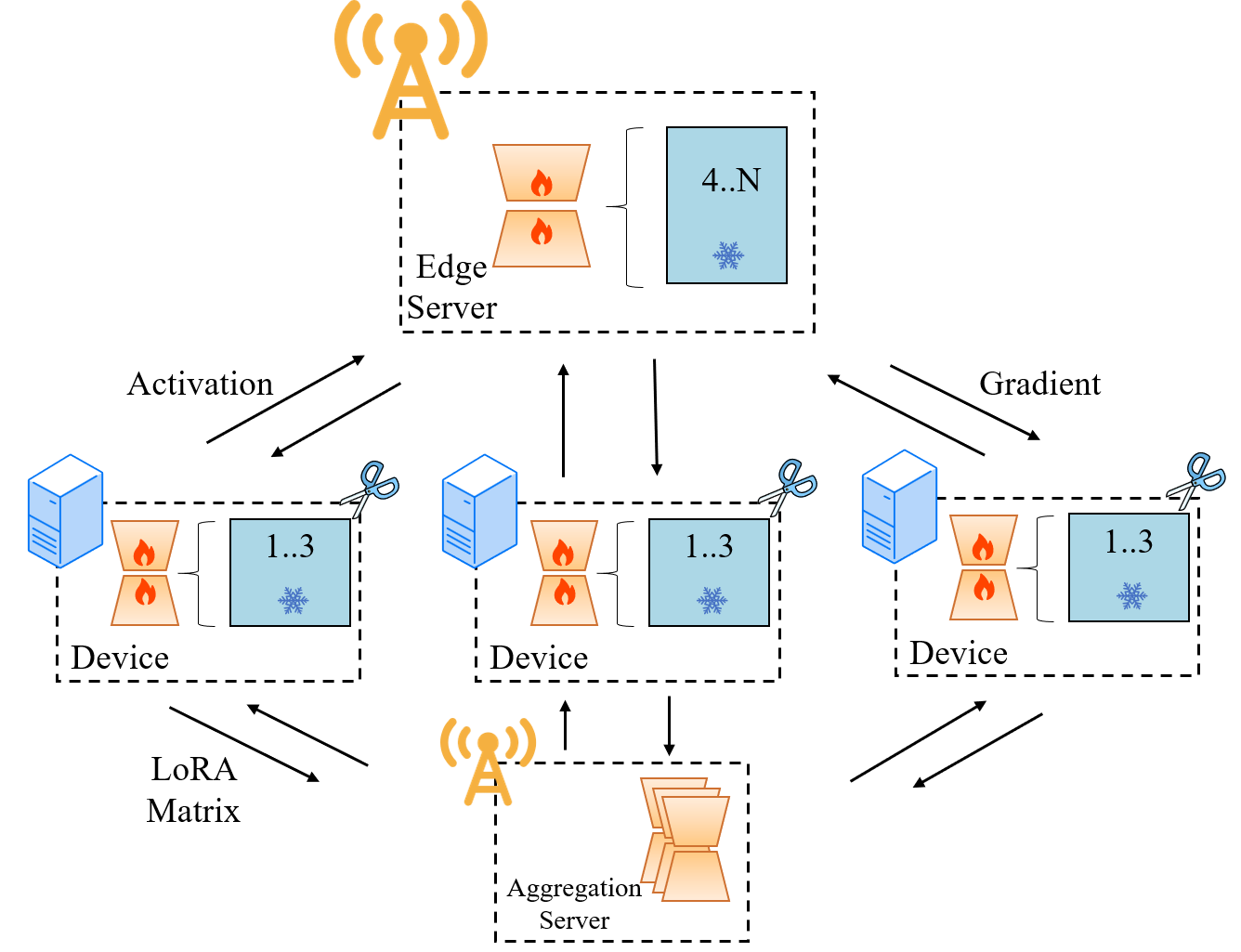}
    \caption*{(b) SplitLoRA[4]} 
    \label{fig:fedlora}
  \end{minipage}
  
  \vspace{10pt}  
  
  \begin{minipage}[b]{\textwidth}
    \centering
    \includegraphics[width=0.7\linewidth]{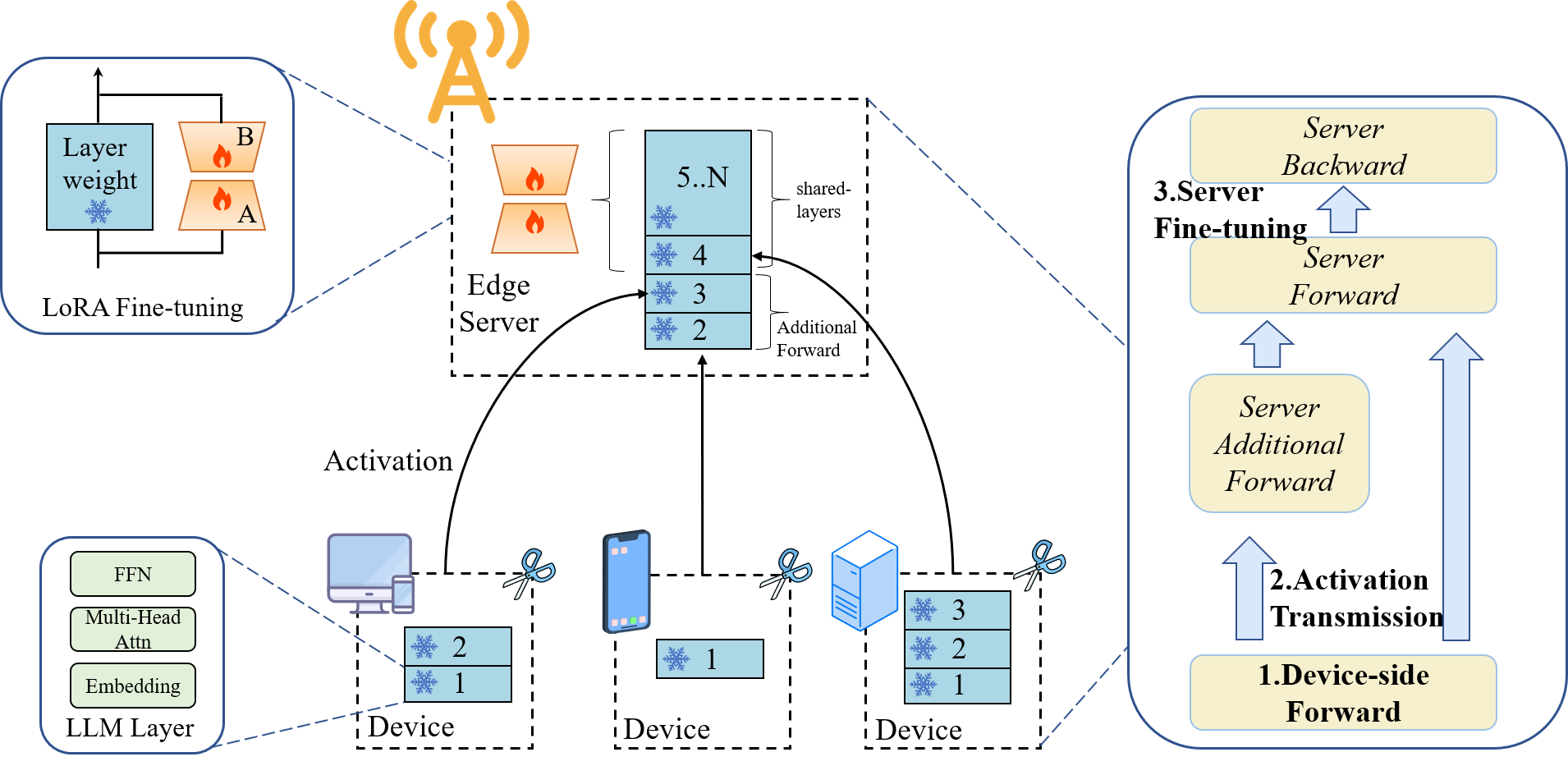}
    \caption*{(c) Proposed SplitFrozen Framework} 
    \label{fig:fedlora}
  \end{minipage}
  
  \caption{Structural diagram of FedLoRA [3], SplitLoRA [4], and SplitFrozen for fine-tuning LLMs on mobile devices.}
  \label{fig:structure}
\end{figure*}

\section{Preliminaries on Deploying LLMs on Resource-Constrained Devices}

Deploying LLMs on resource-constrained devices, such as smartphones, IoT devices, and embedded systems, requires overcoming inherent limitations in computational power, memory, and energy. Two complementary strategies have emerged to address these challenges, i.e., \textit{(1) Model lightweighting,} which optimizes the LLM architecture itself for computation reduction, and \textit{(2) Net4LLM,} which emphasizes on orchestrating the computation and communication resources of networks to facilitate LLM on resource-constrained devices.

\subsection{Model Lightweighting}

Model lightweighting reduces the computational and storage demands of LLM while preserving their core capabilities:  \textit{(1) Distillation} training a compact student model to mimic a larger LLM (e.g., MiniLLM retains 89.4\% of GPT-2’s performance with 77.3\% fewer parameters); \textit{(2) Quantization} reducing weight precision (e.g., GPTQ enables LLAMA-13B to run on 24GB GPUs with \textless 1\% accuracy loss); \textit{(3) Pruning} removing redundant parameters (e.g., SparseGPT prunes 50\% of OPT-175B’s weights, cutting memory usage by 60\%); \textit{(4) Parameter sharing} reusing model parameters to prevent overfitting (e.g., Basis Sharing retains full performance with 50\% shared parameters).

\begin{table*}
\centering
\caption{Preliminaries on deploying LLMs on resource-constrained devices}
\begin{tabular}{|>{\centering\arraybackslash}m{0.14\linewidth}|>{\centering\arraybackslash}m{0.07\linewidth}|>{\centering\arraybackslash}m{0.09\linewidth}|>{\centering\arraybackslash}m{0.55\linewidth}|}
\hline
Research Direction & Category & Literature & Highlight \\ 
\hline
\multirow{4}{*}[-1.6\baselineskip]{\centering Model Lightweighting} 
  & Distillation & MiniLLM & Train a small student model to mimic a larger one, cutting parameters while keeping most performance, achieving efficient compression. \\ 
\cline{2-4}
  & Quantization & GPTQ & Reduce weight precision to lower computational needs while preserving performance, improving deployment feasibility.
 \\ 
\cline{2-4}
  & Pruning & SparseGPT & Remove redundant parameters to optimize the model structure, reducing memory usage and effectively improving computational efficiency. \\ 
\cline{2-4}
  & Parameter Sharing & Basis Sharing & Share model parameters while still retaining most of the performance, thus enabling efficient training in resource-constrained environments. \\ 
\hline
\multirow{10}{*}[-4\baselineskip]{\centering Net4LLM}
  & \multirow{5}{=}[-1.7\baselineskip]{\centering FL4LLM} 
    & Fate-LLM~\cite{fed-01} & The first industrial-grade framework addressing resource constraints and data privacy, enabling small businesses to efficiently adopt LLMs. \\ 
\cline{3-4}
    & & PWFL~\cite{fed-02} & Introduce PFIT and PFTT methods, focusing on personalized fine-tuning in wireless networks to enhance privacy and communication efficiency. \\ 
\cline{3-4}
    & & FLRA~\cite{fed-03} & Combine LoRA and wireless computing to optimize fine-tuning efficiency in wireless networks, reducing computation and communication overhead. \\ 
\cline{3-4}
    & & FFT~\cite{fed-04} & Address deployment challenges due to wireless network resource constraints, ensuring fine-tuning effectiveness and efficiency. \\ 
\cline{3-4}
    & & LLMEA~\cite{fed-05} & Execute AI models at minimal cost and applies AI models on edge servers. \\ 
\cline{2-4}
  & \multirow{5}{=}[-3\baselineskip]{\centering SL4LLM} 
    & Splitwise~\cite{split-01} & Inference into two stages to optimize resources and cut costs. \\ 
\cline{3-4}
    & & SFT~\cite{split-02} & Split LLM across edge servers and mobile devices, managing memory and communication overhead, suitable for wireless networks. \\ 
\cline{3-4}
    & & MobiLLM~\cite{split-03} & Enable memory-efficient fine-tuning on mobile devices with server assistance, reducing computational burden. \\ 
\cline{3-4}
    & & EESL~\cite{split-04} & Optimize energy efficiency in edge network-based LLM fine-tuning. \\ 
\cline{3-4}
    & & Snake Learning~\cite{split-05} & Propose an efficient distributed learning framework for 6G networks, adaptable to heterogeneous environments and suitable for CV and LLM tasks. \\
\cline{3-4}
    & & SplitFrozen (This paper) & The first framework that enables efficient fine-tuning LLMs on heterogeneous resource-constrained devices and imbalanced data, by strategically freezing device-side model layers while centralizing LoRA on the server.\\
\hline
\end{tabular}
\label{tab:your_label}
\end{table*}

\subsection{Network for LLM (Net4LLM)}
Net4LLM focuses on network-centric strategies to optimize the interplay between devices and edge servers to support LLM workflows. The studies on Net4LLM can be categorized into:

\textit{(1) Federated learning for LLM (FL4LLM).} FL offers a privacy-preserving framework for distributed training of LLMs by only exchanging the model parameters. FATE-LLM~\cite{fed-01} introduces an industrial-grade FL framework for LLM fine-tuning, tackling resource constraints and data privacy. PWFL~\cite{fed-02} focuses on wireless networks, proposing personalized federated instruction tuning (PFIT) and personalized federated task tuning (PFTT) to enhance communication efficiency and privacy. FLRA~\cite{fed-03} further integrates LoRA and wireless computation for efficient LLM fine-tuning, reducing computation and communication costs. FFT~\cite{fed-04} optimizes the model deployment for resource-constrained devices, focusing on device heterogeneity and resource allocation in wireless networks. Autonomous edge AI powered by LLMs~\cite{fed-05} extends federated learning to 6G edge AI, offering an efficient edge server execution scheme that reduces costs and boosts model performance.

\textit{(2) Split learning for LLM (SL4LLM).} By partitioning and offloading model portions to the edge server, split learning (SL) is well-suited for fine-tuning LLM on resource-constrained devices. Splitwise~\cite{split-01} divides LLM inference into prompt computation and token generation phases, optimizing resource allocations via split learning. Split Fine-Tuning (SFT)~\cite{split-02} applies  the concept of Splitwise to LLM fine-tuning in wireless networks,  mitigating the bottleneck of device memory and communication overhead. MobiLLM~\cite{split-03} designs efficient LLM fine-tuning framework on resource-constrained devices with server-assisted side-tuning. Energy-Efficient Split Learning (EESL)~\cite{split-04} proposes an energy-efficient SL framework for edge networks, optimizing latency and energy consumption while accounting for device heterogeneity. Snake learning~\cite{split-05} proposes a 6G distributed learning framework with hierarchical serpentine updates, adaptable to LLM tasks.

However, none of the existing studies (as shown in Table I) can address the challenges arising from device heterogeneity and data imbalance. This paper is the first framework to enable efficient fine-tuning LLMs on heterogeneous resource-constrained devices and imbalanced data.

\section{Proposed SplitFrozen Framework}
This section presents the proposed SplitFrozen framework that enables efficient fine-tuning of LLMs by integrating layer-wise model partitioning, device-side frozen layers, and server-centric LoRA. SplitFrozen addresses the challenges of device heterogeneity and data imbalance while minimizing computational overhead on resource-constrained devices.  The details are as follows.

\subsection{Overview of SplitFrozen Framework}

\textbf{System Architecture.} As shown in Fig. 1(c), SplitFrozen operates through a collaborative device-server architecture: \textit{(1) Device-side Forward.} Resource-constrained devices deploy a subset of the model’s frozen layers (e.g., the first 1–3 layers of a 12-layer GPT-2). These devices perform forward propagation only, eliminating backpropagation and gradient computation overhead. \textit{(2) Activation Transmission.} Device-sides transmit the generated activations to the server. \textit{(3) Server Fine-tuning.} The server receives the activations transmitted from the device-sides, executes forward/backward propagation of the remaining layers, and updates parameters efficiently via LoRA. 

The framework allows the differently partitioned device-side model portions (i.e., with different layers of transformers). The server with a global model can help conduct the forward propagation for the devices with small model portions to the same layer with others,  and then perform LoRA update of the remaining layers. This design dynamically accommodates device heterogeneity by deploying different numbers of layers while centralizing resource-intensive tasks on the server, and it leverages a three-stage pipeline to efficiently speed up the model fine-tuning process across devices and the server.
 
\textbf{Training Workflow.} The training process of SplitFrozen consists of three steps, i.e., Device-side Forward, Activation Transmission and Server Fine-tuning:

\textit{(1) Device-side Forward.} Devices deploy the first few layers of the model based on its computational capacity (e.g., for a 12-layer GPT-2 model, a device may deploy the first 1 to 3 layers). Given locally available task-specific data, the device performs forward propagation, generating activations. These activations serve as inputs for subsequent computations, while no local weight updates are required.No weight updates or gradients are computed on-device, eliminating the expensive overhead of backpropagation and gradient computation. The device-side frozen strategy is particularly well-suited for resource-constrained devices, ensuring that even low-computation devices can participate in the training process.

\textit{(2) Activation Transmission.} Devices transmit activations to the server. Activation size remains constant across devices (e.g., 6MB for GPT-2) due to standardized transformer layer dimensions. Compression techniques in model lightweighting (e.g., quantization and pruning) can be integrated to further reduce the communication overhead.

\textit{(3) Server Fine-tuning.} The server is responsible for receiving activations from devices and using them as the starting point for subsequent computations. To handle heterogeneous device capability (different local model layers), the devices are also required to transmit a layer-count metadata to the server (only once at the initialization when partitioning the model). When fewer layers are deployed on the device-sides, they will connect to the lower layers on the server, resulting in additional computations (e.g., Additional Forward on the server in Fig. 1(c)). Subsequently, the server's shared layers undergo LoRA fine-tuning. Note that, after aggregating and shuffling the activations of all devices, the server can perform centralized LoRA to improve model accuracy under Non-IID data distributions.

\textbf{Summary.} The proposed SplitFrozen framework addresses the challenges of heterogeneous resource-constrained devices and imbalanced data via: \textit{(1) Differently Device-side Model Frozen.} Devices dynamically deploy frozen layers based on hardware capabilities (only forward propagation), while the server adapts to variable input depths, ensuring that even low-power devices (e.g., Raspberry Pi) can participate without throttling the system; \textit{(2) Aggregated Server-centric LoRA.} The server can perform centralized LoRA by shuffling the device activations to eliminate Non-IID degradation.

\subsection{Pipeline Parallelism for Wireless Fine-tuning}

SplitFrozen proposes a pipeline parallelism mechanism to optimize the processes of the device-side forward propagation, activation transformation, and server-side fine-tuning, thereby accelerating the fine-tuning process.

\begin{figure}
    \centering
    \includegraphics[width=1\linewidth]{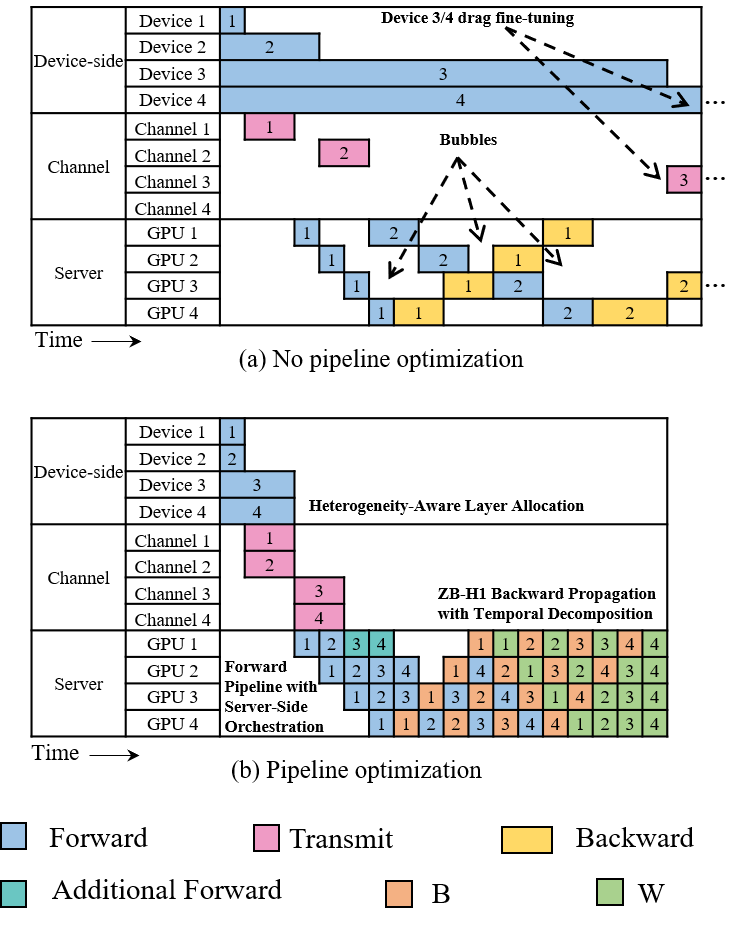}
    \caption{ The pipelining parallelism optimization of SplitFrozen to accelerate wireless fine-tuning.}
    \label{fig:enter-label}
\end{figure}
 
\textit{(1) Heterogeneity-Aware Layer Allocation.} The computational heterogeneity of device-side nodes leads to imbalanced forward computation and activation transmission times (if applying the same LLM partitioning decisions), resulting in suboptimal pipeline efficiency. For example, Fig. 2(a) shows prolonged fine-tuning latency when computationally weak devices, e.g., Device 3/4, are assigned the same number of LLM layers. To address this, SplitFrozen strategically partitions the LLM model portions by allocating more LLM layers to the devices with higher computational capabilities, e.g., Devices 1/2 in Fig. 2(b). This ensures synchronized forward propagation and activation transmission timelines across heterogeneous devices, enabling parallel workflow optimization.

\textit{(2) Forward Pipeline with Server-Side Orchestration.} Traditional forward pipelines, which offload partial computations to servers, often suffer from temporal misalignment between server-side forward/backward passes and device-side operations, creating idle intervals, i.e., time bubbles, as shown in Fig. 2(a). SplitFrozen mitigates this by decoupling server-side tasks: the server first completes additional forward propagation for devices with fewer layers, e.g., Device 3/4 with limited resources in Fig. 2(b), before initiating shared-layer forward/backward processing. This sequencing minimizes synchronization overhead and eliminates pipeline stalls.

\textit{(3) ZB-H1 Backward Propagation with Temporal Decomposition.} Building on the ZB-H1 algorithm, SplitFrozen refines backward propagation into two fine-grained temporal phases: gradient computation ($B$) and weight update ($W$), with the constraint $\textit{Backward} = B + W$, where $B$ must be completed before $W$. By interleaving and overlapping $B$ and $W$ computations, SplitFrozen achieves higher pipeline utilization. Moreover, we holistically consider the computation speeds and timing of the device-side forward, channel activation transmission, and server fine-tuning, ensuring that these fine-grained time segments coordinate effectively with minimal discrepancy.

\section{Experimental Results}

\subsection{Experimental setup}
{\bf{Hardware and Software Environment:}} Experiments were conducted on four NVIDIA GeForce RTX 4090 GPUs (24GB VRAM per GPU) using Python 3.12 and PyTorch 2.0.1.

{\bf{Datasets and Partitioning:}} We evaluated SplitFrozen on four benchmarks: (1) MRPC, an NLP dataset for paraphrase identification (3668 train, 408 validation, 1725 test examples). (2) MNLI-matched, an NLI dataset (9815 validation examples) assessing entailment, contradiction, or neutrality in premise-hypothesis pairs. (3) SST-2, an NLP dataset for binary sentiment classification (67315 train, 872 validation, 1821 test examples). (4) GSM8K, a math reasoning dataset (7500 train, 1000 test examples) for multi-step problem-solving. Since test answers for MRPC, MNLI-matched, and SST-2 are unavailable, we use MRPC and SST-2 validation sets as test sets. For MNLI-matched, 10\% of the data is randomly selected as the test set, with the rest for training. We adopt both IID and Non-IID data distributions to partition the data across different devices. Dirichlet distribution (with configurable concentration parameter $\alpha$) is used to generate the imbalanced data.

{\bf{Model:}}We used the GPT-2 small for NLP experiments and the Llama-3.2-1B specifically for the Mathematical Reasoning test. GPT-2 small is a large language generation model developed by OpenAI, with 124 million parameters and 12 transformer layers.Llama-3.2-1B is a large language generation model developed by Meta, with 1 billion parameters and 16 transformer layers.

\textbf{Hyperparameter.} The sequence length is 128 tokens for MRPC/SST-2, and 256 tokens for MNLI-matched/GSM8K. The batch size and learning rate are 72 and 5e-5, respectively. The rank of LoRA is \( R = \{4, 8\} \). 

{\bf{Experimental Setup: }}In our simulation, we deployed one server and ten devices. The server has a peak performance of 330.4 TFLOPS, equivalent to four NVIDIA RTX 4090 GPUs. To ensure that the three stages of the pipeline have approximately equal time granularities, the simulation is configured as follows. The ten devices vary in computational capability, corresponding to 10\% and 20\% of the peak performance of a single NVIDIA RTX 4090, with 3 devices at the 10\% level and 7 devices at the 20\% level. We deploy the first 1 layer and 3 layers of GPT-2 small or Llama-3.2-1B on the devices with 10\% and 20\% computational capacity, respectively, while the remaining 11 layers (for GPT-2 small) or 15 layers (for Llama-3.2-1B) are deployed on the server. The communication channel between the devices and the server is set to a transmission rate of 600 Mbps.

\begin{table*}[]
\caption{Comparison of model accuracy, computational overhead (in terns of MFLOPs), and running time of different approaches.}
\label{tab:comparison}
\resizebox{\textwidth}{!}{
\begin{tabular}{|c|c|ccccc|ccccc|ccccc|}
\hline
{\color[HTML]{404040} } &  & \multicolumn{5}{c|}{MRPC} & \multicolumn{5}{c|}{MNLI-matched} & \multicolumn{5}{c|}{SST2} \\ \cline{3-17} 
\multirow{-2}{*}{{\color[HTML]{404040} Rank}} & \multirow{-2}{*}{Method} & IID & \makecell{Non-IID \\ $\left( \alpha = 0.1 \right)$} & \makecell{Device \\ MFLOP} & \makecell{Device \\ Time} & \makecell{Total \\ Time} & IID & \makecell{Non-IID \\ $\left( \alpha = 0.1 \right)$} & \makecell{Device \\ MFLOP} & \makecell{Device \\ Time} & \makecell{Total \\ Time} & IID & \makecell{Non-IID \\ $\left( \alpha = 0.1 \right)$} & \makecell{Device \\ MFLOP} & \makecell{Device \\ Time} & \makecell{Total \\ Time} \\ \hline
{\color[HTML]{404040} } & CenLoRA & 74.0 & \textbf{74.0} & 513 & \cellcolor[HTML]{EFEFEF}19.8 & \cellcolor[HTML]{EFEFEF}19.8 & 63.6 & 63.6 & 1375 & \cellcolor[HTML]{EFEFEF}45.8 & \cellcolor[HTML]{EFEFEF}45.8 & \textbf{86.2} & \textbf{86.2} & 9410 & \cellcolor[HTML]{EFEFEF}221.7 & \cellcolor[HTML]{EFEFEF}221.7 \\
{\color[HTML]{404040} } & FedLoRA~\cite{reffedlora} & 64.3 & 45.7 & 513 & \cellcolor[HTML]{EFEFEF}19.5 & \cellcolor[HTML]{EFEFEF}19.8 & 34.9 & 35.8 & 1375 & \cellcolor[HTML]{EFEFEF}45.5 & \cellcolor[HTML]{EFEFEF}46.0 & 79.4 & 34.4 & 9410 & \cellcolor[HTML]{EFEFEF}222.4 & \cellcolor[HTML]{EFEFEF}222.7 \\
{\color[HTML]{404040} } & SplitLoRA~\cite{refsplitlora} & 70.8 & 55.5 & 128 & \cellcolor[HTML]{EFEFEF}22.7 & \cellcolor[HTML]{EFEFEF}30.2 & 60.6 & 32.3 & 344 & \cellcolor[HTML]{EFEFEF}78.2 & \cellcolor[HTML]{EFEFEF}104.2 & 79.1 & 56.4 & 2353 & \cellcolor[HTML]{EFEFEF}360.3 & \cellcolor[HTML]{EFEFEF}480.4 \\
\multirow{-4}{*}{{\color[HTML]{404040} 4}} & SplitFrozen & \textbf{76.6} & 73.2 & \textbf{43} & \cellcolor[HTML]{EFEFEF}\textbf{7.7} & \cellcolor[HTML]{EFEFEF}\textbf{9.7} & \textbf{65.4} & \textbf{66.3} & \textbf{114} & \cellcolor[HTML]{EFEFEF}\textbf{36.6} & \cellcolor[HTML]{EFEFEF}\textbf{45.7} & 79.4 & 80.3 & \textbf{779} & \cellcolor[HTML]{EFEFEF}\textbf{139.6} & \cellcolor[HTML]{EFEFEF}\textbf{170.3} \\ \hline
 & CenLoRA & 76.2 & \textbf{76.2} & 517 & \cellcolor[HTML]{EFEFEF}19.9 & \cellcolor[HTML]{EFEFEF}19.9 & 63.0 & \textbf{63.0} & 1385 & \cellcolor[HTML]{EFEFEF}46.2 & \cellcolor[HTML]{EFEFEF}46.2 & \textbf{88.8} & \textbf{88.8} & 9476 & \cellcolor[HTML]{EFEFEF}223.3 & \cellcolor[HTML]{EFEFEF}223.3 \\
 & FedLoRA~\cite{reffedlora} & 63.3 & 43.9 & 517 & \cellcolor[HTML]{EFEFEF}19.5 & \cellcolor[HTML]{EFEFEF}20.0 & 41.7 & 35.8 & 1385 & \cellcolor[HTML]{EFEFEF}45.8 & \cellcolor[HTML]{EFEFEF}46.3 & 78.9 & 34.4 & 9476 & \cellcolor[HTML]{EFEFEF}223.6 & \cellcolor[HTML]{EFEFEF}224.2 \\
 & SplitLoRA~\cite{refsplitlora} & 75.2 & 55.5 & 129 & \cellcolor[HTML]{EFEFEF}22.8 & \cellcolor[HTML]{EFEFEF}30.4 & 57.1 & 32.3 & 346 & \cellcolor[HTML]{EFEFEF}78.7 & \cellcolor[HTML]{EFEFEF}104.9 & 80.5 & 53.8 & 2370 & \cellcolor[HTML]{EFEFEF}362.9 & \cellcolor[HTML]{EFEFEF}483.8 \\
\multirow{-4}{*}{8} & SplitFrozen & \textbf{78.7} & 74.6 & \textbf{43} & \cellcolor[HTML]{EFEFEF}\textbf{7.7} & \cellcolor[HTML]{EFEFEF}\textbf{9.7} & \textbf{65.2} & 61.7 & \textbf{114} & \cellcolor[HTML]{EFEFEF}\textbf{36.6} & \cellcolor[HTML]{EFEFEF}\textbf{45.7} & 80.6 & 80.7 & \textbf{779} & \cellcolor[HTML]{EFEFEF}\textbf{139.6} & \cellcolor[HTML]{EFEFEF}\textbf{170.3} \\ \hline
\end{tabular}
}
\end{table*}

\begin{figure}
    \centering
    \includegraphics[width=1\linewidth]{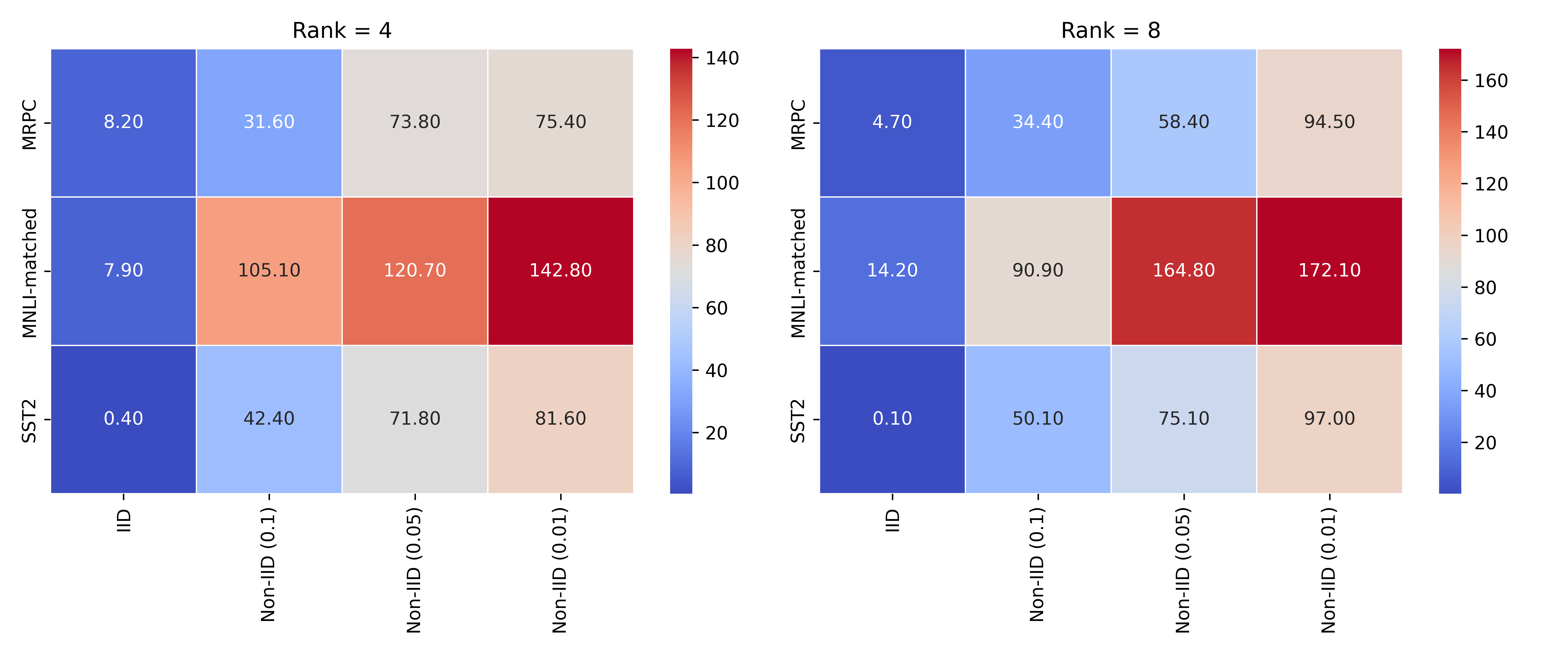}
    \caption{The performance gain of SplitFrozen over SplitLoRA~\cite{refsplitlora} under different matrix ranks and data distributions.}
    \label{fig:enter-label}
\end{figure}

\begin{table*}[t]
    \caption{Comparison of fine-tuning Llama-3.2-1B via different approaches on math question-answer generation.}
    \label{fig:comparison_table}
    \centering
    \renewcommand{\arraystretch}{1.2}
    \begin{tabular}{|>{\centering\arraybackslash}m{0.15\linewidth}|p{0.75\linewidth}|}
        \hline
        Question & Tina makes \$18.00 an hour. If she works more than 8 hours per shift, she is eligible for overtime, which is paid by your hourly wage + 1/2 your hourly wage. If she works 10 hours every day for 5 days, how much money does she make? \\ \hline
        CenLoRA & Tina earns \$18.00 per hour. For the first 8 hours, she makes \$144.00. For the 2 overtime hours, she earns \$27.00 per hour. So, she makes \$54.00 in overtime. Total daily pay is \$198.00. Over 5 days, she earns \$990.00. \\ \hline
        FedLoRA[3] & Tina works 10 hours a day, so she makes a lot. She probably earns over \$900 for the week. \\ \hline
        SplitLoRA[4] & Tina’s regular pay is \$18.00 per hour. She works 10 hours a day and gets overtime, so she likely makes around \$800 for 5 days. \\ \hline
        SplitFrozen & Tina’s hourly rate is \$18.00. She works 10 hours a day, so she makes \$144.00 for the first 8 hours and \$54.00 for 2 overtime hours. Total daily pay is \$198.00, and for 5 days, her total earnings are \$990.00. \\ \hline
        Question & There are 25 roses in a garden. There are 40 tulips. There are 35 daisies. What percentage of flowers are not roses? \\ \hline
        CenLoRA & There are 25 roses, 40 tulips, and 35 daisies, totaling 100 flowers. The number of flowers that are not roses is 75. So, 75\% of the flowers are not roses. \\ \hline
        FedLoRA[3] & There are 100 flowers in total, and most of them are not roses. So, a big percentage of the flowers are not roses. \\ \hline
        SplitLoRA[4] & There are 25 roses, so probably most of the flowers are not roses. It's around 75\%. \\ \hline
        SplitFrozen & Out of 100 flowers (25 roses, 40 tulips, 35 daisies), 75 are not roses. That means 75\% of the flowers are not roses. \\ \hline
    \end{tabular}
\end{table*}

\subsection{Results Analysis}

\subsubsection{Fine-tuning Accuracy}

Table \ref{tab:comparison} compares the fine-tuning performance of SplitFrozen with other baseline algorithms using the GPT-2 small model, with accuracy as the performance metric. Under the IID distribution, SplitFrozen achieves an average accuracy improvement of 13.6\% over FedLoRA and SplitLoRA. Under the Non-IID distribution, SplitFrozen outperforms FedLoRA and SplitLoRA by an average of 69.4\%. In Fig. 3, we further explore scenarios with more imbalanced data distributions and find that SplitFrozen surpasses SplitLoRA by up to 172.1\%. In some cases, SplitFrozen even outperforms CenLoRA. This is because lower transformer layers primarily learn general features, and freezing the initial layers enhances the generalization capability of the LLM. SplitFrozen adopts a centralized training approach on the server, effectively addressing the heterogeneity among different devices.

\subsubsection{Parallel analysis}

Table \ref{tab:comparison} compares the computational load and fine-tuning time differences of various algorithms in terms of Device MFLOP, Device Time, and Total Time. On average, SplitFrozen reduces device computation and fine-tuning time by 86.8\% and 51.0\% compared to FedLoRA and SplitLoRA, respectively.Additionally, SplitFrozen reduces the total fine-tuning time by 50.2\% compared to FedLoRA and SplitLoRA. This is because SplitFrozen only performs forward propagation on the device, eliminating the need to compute gradients. Moreover, the device can deploy fewer layers, significantly reducing computational load and time. Furthermore, SplitFrozen utilizes a device-channel-server pipeline design to further decrease the total fine-tuning time.

\subsubsection{Content-generation LLM Test}

In this study, we evaluated the performance of four different algorithms using the Llama-3.2-1B model on the GSM8K dataset. In Table \ref{fig:comparison_table}, our results indicate that SplitFrozen demonstrates performance comparable to centralized algorithms, outperforming both FedLoRA and SplitLoRA. While FedLoRA and SplitLoRA showed some promise, their capabilities were notably inferior to SplitFrozen, which exhibited more consistent and accurate results across a variety of test cases. These findings suggest that SplitFrozen is a highly competitive alternative, particularly when compared to the other federated learning approaches tested in this study.

\section{Conclusion}
This article presents SplitFrozen, a distributed framework designed to address the challenges of fine-tuning LLMs on heterogeneous resource-constrained devices and Non-IID data distribution. By strategically partitioning LLMs into device-side frozen portion and server LoRA portion, SplitFrozen drastically reduces computational demands on devices while maintaining high accuracy. Experimental results on GPT-2 and Llama-3.2 demonstrate the superiority of SplitFrozen over state-of-the-art frameworks, outperforming by 69.4\% in accuracy under Non-IID distributions while reducing up to 86.8\% device-side computations and 50.2\% total training time. 

SplitFrozen bridges the gap between massive LLMs and resource-constrained device deployment, offering a practical solution for privacy-sensitive, resource-efficient AI applications in next-generation networks. The future directions may include (1) integration of activation compression to reduce communication overhead and (2) enhancing scalability for 6G networks with multiple servers, fostering deeper AI-communication convergence.


\begin{thebibliography}{99}

\bibitem{ref1}
S.~Alikhani, G.~Charan, and A.~Alkhateeb, ``Large Wireless Model ({LWM}): A Foundation Model for Wireless Channels,'' \emph{arXiv:2411.08872}, 2024.

\bibitem{ref2}
Z.~Lin, G.~Qu, Q.~Chen, \emph{et al.}, ``Pushing Large Language Models to the {6G} Edge: Vision, Challenges, and Opportunities,'' \emph{arXiv:2309.16739}, 2023.

\bibitem{reffedlora}
X.~Wu, X.~Liu, J.~Niu, \emph{et al.}, ``{FedLoRA}: When Personalized Federated Learning Meets Low-Rank Adaptation,'' in \emph{Proc. ICLR}, 2024.

\bibitem{refsplitlora}
Z.~Lin, X.~Hu, Y.~Zhang, \emph{et al.}, ``{SplitLoRA}: A Split Parameter-Efficient Fine-Tuning Framework for Large Language Models,'' \emph{arXiv:2407.00952}, 2024.

\bibitem{splitfed}
C.~Thapa, P.~C.~M.~Arachchige, S.~Camtepe, \emph{et al.}, ``{SplitFed}: When Federated Learning Meets Split Learning,'' in \emph{Proc. AAAI}, vol.~36, no.~8, pp. 8485--8493, 2022.

\bibitem{fed-01}
T.~Fan, Y.~Kang, G.~Ma, \emph{et al.}, ``{FATE-LLM}: An Industrial Grade Federated Learning Framework for Large Language Models,'' \emph{arXiv:2310.10049}, 2023.

\bibitem{fed-02}
F.~Jiang, L.~Dong, S.~Tu, \emph{et al.}, ``Personalized Wireless Federated Learning for Large Language Models,'' \emph{arXiv:2404.13238}, 2024.

\bibitem{fed-03}
H.~Sun, H.~Tian, W.~Ni, \emph{et al.}, ``Federated Low-Rank Adaptation for Large Models Fine-Tuning over Wireless Networks,'' \emph{IEEE Trans. Wireless Commun.}, 2024.

\bibitem{fed-04}
Z.~Wang, Y.~Zhou, Y.~Shi, \emph{et al.}, ``Federated Fine-Tuning for Pre-Trained Foundation Models Over Wireless Networks,'' \emph{IEEE Trans. Wireless Commun.}, 2025.

\bibitem{fed-05}
Y.~Shen, J.~Shao, X.~Zhang, \emph{et al.}, ``LLMs Empowered Autonomous Edge {AI} for Connected Intelligence,'' \emph{IEEE Commun. Mag.}, vol.~62, no.~10, pp. 140--146, 2024.

\bibitem{split-01}
P.~Patel, E.~Choukse, C.~Zhang, \emph{et al.}, ``{SplitWise}: Efficient Generative {LLM} Inference Using Phase Splitting,'' in \emph{Proc. ISCA}, pp. 118--132, 2024.

\bibitem{split-02}
S.~Zhang, G.~Cheng, X.~Huang, \emph{et al.}, ``Split Fine-Tuning for LLMs in Wireless Networks,'' \emph{arXiv:2501.09237}, 2025.

\bibitem{split-03}
L.~Li, X.~Yang, W.~Wu, \emph{et al.}, ``{MobiLLM}: Enabling {LLM} Fine-Tuning on Mobile Devices via Server-Assisted Side Tuning,'' \emph{arXiv:2502.20421}, 2025.

\bibitem{split-04}
Z.~Li, S.~Wu, L.~Li, \emph{et al.}, ``Energy-Efficient Split Learning for LLM Fine-Tuning in Edge Networks,'' \emph{IEEE Netw. Lett.}, 2025.

\bibitem{split-05}
X.~Yu, X.~Yi, R.~Li, \emph{et al.}, ``Snake Learning: A Communication- and Computation-Efficient Distributed Learning Framework for {6G},'' \emph{IEEE Commun. Mag.}, 2025.


\end{thebibliography}
\end{document}